# Underwater Image Super-Resolution using Generative Adversarial Network-based Model

Alireza Aghelan, Modjtaba Rouhani
Computer Engineering Department, Ferdowsi University of Mashhad
Mashhad, Iran
aghelan.alireza@mail.um.ac.ir, rouhani@um.ac.ir

*Abstract*—Single image super-resolution (SISR) models are able to enhance the resolution and visual quality of underwater images and contribute to a better understanding of underwater environments. The integration of these models in Autonomous Underwater Vehicles (AUVs) can improve their performance in vision-based tasks. Real-Enhanced Super-Resolution Generative Adversarial Network (Real-ESRGAN) is an efficient model that has shown remarkable performance among SISR models. In this paper, we fine-tune the pre-trained Real-ESRGAN model for underwater image super-resolution. To fine-tune and evaluate the performance of the model, we use the USR-248 dataset. The fine-tuned model produces more realistic images with better visual quality compared to the Real-ESRGAN model.

## I. Introduction

Autonomous Underwater Vehicles (AUVs) are used in important applications including marine species monitoring, human-robot collaboration, underwater scene analysis, seabed mapping, and others. [1] These vehicles need good-quality images to effectively perform their tasks such as tracking, scene understanding, etc. By using precise visual information, AUVs can make informed decisions, navigate complex underwater environments, and contribute to the success of diverse underwater missions.

Underwater images are often degraded for various reasons, including absorption, scattering, and poor visibility. These degradations result in the loss of important details and blurred images. Despite the use of advanced cameras, this problem remains unresolved, making it difficult to process underwater images. Fast and accurate image super-resolution methods can help in addressing this problem and improve the resolution and visual quality of underwater images. [1]

One of the significant challenges in the field of underwater image super-resolution is the lack of datasets containing high-quality and high-resolution (HR) images. Obtaining such datasets is challenging due to the difficulties associated with capturing clear and detailed underwater scenes. However, in this work, we address this challenge by using the USR-248 dataset, which comprises a collection of good-quality underwater images. By utilizing this dataset, we can enhance the robustness and effectiveness of our proposed method in producing HR outputs with improved visual quality.

Another important challenge is the limited amount of data in most underwater datasets. These datasets contain only a few hundred images, which is often insufficient for models to learn the intricate patterns in underwater scenes. This limitation makes it difficult to train complex models from scratch. To overcome this challenge, we use transfer learning technique. By adopting this approach, we leverage existing knowledge from pre-trained models to improve our models' performance.

In dynamic underwater environments, the speed of image processing can greatly impact an AUV's ability to perform tasks such as obstacle avoidance, navigation, and rapid decision-making. Therefore, it is important to note that the underwater image super-resolution models should have high computational efficiency and rapid inference time. This ensures that AUVs utilizing these models can deliver efficient and reliable performance when processing underwater images, enabling them to effectively perform their tasks.

Traditional single image super-resolution (SISR) methods, such as interpolation-based and reconstruction-based methods, do not perform well in recovering high-frequency information. They do not consider the complex non-linear mapping between low and high-resolution images. These methods usually produce blurry images that lack fine textures. Nowadays, deep learning-based SISR methods are widely used and outperform most traditional methods. These methods can learn to infer photo-realistic details from data rather than relying on mathematical models. Generative Adversarial Network (GAN) [2] based methods have shown substantial improvement among deep learning-based SISR methods, and many studies have been conducted on them.

GAN-based super-resolution models utilize generator and discriminator networks in an adversarial training process. The generator network aims to produce HR outputs indistinguishable from real images. Simultaneously, the discriminator distinguishes between the real images and the generated samples, providing valuable feedback to the generator. This adversarial competition drives the generator to synthesize realistic texture details, resulting in a significant improvement in both visual realism and overall quality of the HR images compared to previous approaches.

Real-Enhanced Super-Resolution Generative Adversarial Network [3] (Real-ESRGAN) is a practical model that is widely used for general image super-resolution. This model has shown excellent performance in recovering HR images from real-world low-resolution (LR) images. Furthermore, the computations in this model are performed at an acceptable speed. These reasons motivated us to fine-tune the pre-trained Real-ESRGAN model on an underwater image dataset and use it to improve the resolution and visual quality of the underwater images.

## II. RELATED WORK

This section provides an overview of several works in the field of underwater image super-resolution alongside a summary description of the Real-ESRGAN model.

Scattering is one of the reasons for the degradation of underwater images. Lu et al. [4] introduced a self-similarity-based approach for de-scattering and enhancing the resolution of underwater images. In traditional methods, most of the high-frequency information is lost during the de-scattering process. The authors addressed this issue by presenting a high turbidity underwater image super-resolution method. The proposed model produces pleasant outputs with a reasonable noise level.

Islam et al. [1] developed SRDRM and SRDRM-GAN models to efficiently perform underwater image super-resolution for use in autonomous underwater robots. The SRDRM is a deep residual network-based generative model. In the SRDRM-GAN model, the SRDRM model performed the role of the generator, and a Markovian PatchGAN-based model operated as the discriminator. The authors introduced an objective function to supervise the training. This function assesses the perceptual quality of an image with respect to its global content, color, and texture information. They also presented the USR-248 dataset that we use in our work. Details about the USR-248 dataset can be found in Section 3.

AlphaSRGAN [5] is a practical GAN-based underwater image super-resolution model that merges traditional image reconstruction approaches with deep learning methods. The authors incorporate pre-processing images before entering them into the generator network, resulting in improved performance and stability. In their work, the USR 248 dataset [1] is utilized for training and evaluation. This model demonstrates remarkable performance and is well-suited for real-time applications.

In Wang et al. [6] the authors proposed a lightweight multi-stage information distillation network to balance computational speed and model performance in underwater image super-resolution tasks. They presented a recursive residual feature distillation module to extract useful features with a small number of parameters. Moreover, a channel interaction & distillation module was used to extract useful information without additional parameters. These modules are very useful in reducing the consumption of computing resources.

Based on the successful prior use of GANs in related works like AlphaSRGAN and SRDRM-GAN, we are motivated to leverage the strengths of generative adversarial networks for underwater image super-resolution. GANs have shown remarkable performance in generating realistic images while improving perceptual quality. Inspired by these advancements, we adopt a state-of-the-art GAN-based super-resolution model called Real-ESRGAN in our work.

Real-ESRGAN [3] model can effectively improve the resolution and visual quality of real-world LR images. This model uses a high-order degradation process to better simulate complicated and real-world degradations. The authors employed a U-Net based discriminator [7] with skip connections to increase discrimination power. Furthermore, they used spectral normalization [8] to stabilize the training process.

In this work, the pre-trained Real-ESRGAN model is fine-tuned for underwater image super-resolution.

## III. METHODOLOGY

In this section, we first present an overview of the dataset used in this paper. Then we explain the methods used to generate LR images and fine-tune the model. Finally, we describe the details of the fine-tuning process.

We utilize the USR-248 dataset [1] for fine-tuning and testing the model. The USR-248 dataset is specifically used for underwater image super-resolution. In this dataset, the training and testing folders each contain multiple sets of LR images paired with the corresponding set of HR images. Each set in the training folder comprises 1060 images, and each set in the testing folder contains 248 images. The original HR images have a resolution of 640 × 480. The USR-248 dataset offers a diverse range of underwater scenes and objects of interest, including various backgrounds and subjects such as coral reefs, fish, divers, and wrecks, and provides a comprehensive collection for training and evaluation purposes.

To fine-tune the model, we use HR images from the training folder. We employ the high-order degradation process of the Real-ESRGAN model to generate LR images from the HR images. The details of the degradation process will be explained in the subsequent paragraphs. To evaluate the performance of the model, we use HR and 4× downsampled images from the testing folder.

In underwater imaging, degradation factors like absorption, scattering, and poor visibility affect image quality. Classical degradation methods are insufficient for simulating these real-world underwater degradations. To address this limitation, it becomes essential to employ a suitable degradation process that better simulates real-world degradations.

The high-order degradation process employed in our research is based on the work of Wang et al. [3]. This process is highly effective for simulating realistic degradations. Employing the high-order degradation process can also be useful for simulating real-world underwater degradations. By integrating this degradation model, we achieve substantial improvements in preserving fine details, reducing noise, and eliminating artifacts. These advancements result in superior perceptual quality and improved performance in underwater image super-resolution tasks.

Transfer learning technique offers several advantages in the context of model training. This technique is useful in compensating for the shortage of data and improves the performance of the model. Moreover, transfer learning facilitates faster convergence of the model and improves generalization capability to unseen data. These benefits make transfer learning a valuable technique for achieving superior performance and efficiency in various deep-learning tasks.

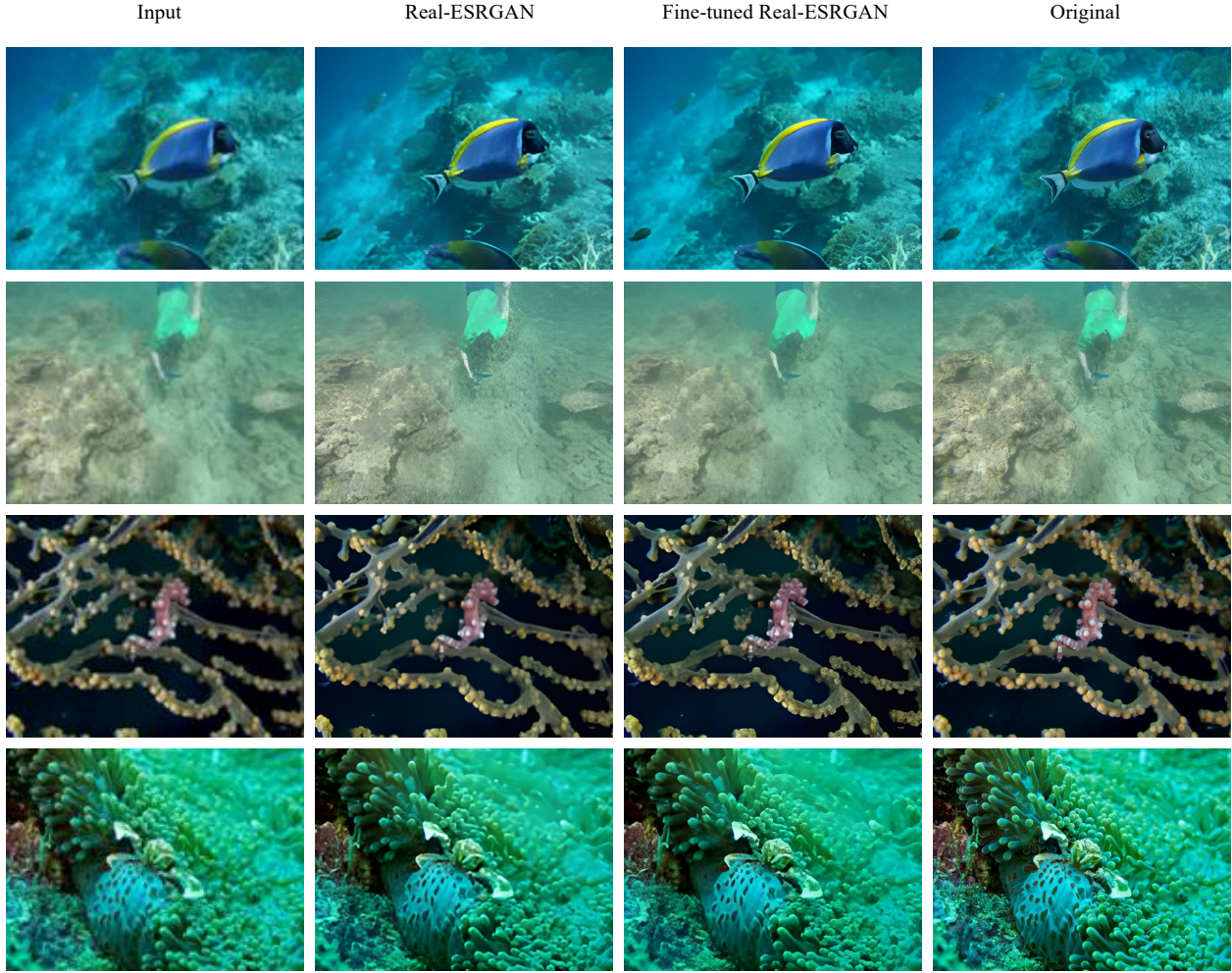

Fig. 1. Columns 1–4 are input images, Real-ESRGAN outputs, fine-tuned Real-ESRGAN outputs, and original images.

In our proposed approach, instead of training the Real-ESRGAN model from scratch, we use transfer learning technique. In addition to the advantages mentioned in the previous paragraph, employing this technique enables us to save significant time and computational resources that are required to train the Real-ESRGAN model from scratch. First, we load the pre-trained generator (v0.1.0) and discriminator (v0.2.2.3) networks of the RealESRGAN_x4plus model from the GitHub repository. This model has been trained on natural image datasets. Next, we fine-tune this model on the USR-248 dataset of underwater images.

In this work, Adam optimizer is used for fine-tuning the pre-trained generator and discriminator networks of the Real-ESRGAN model. We employ a combination of L1 loss, perceptual loss, and GAN loss functions for fine-tuning. The pre-trained Real-ESRGAN model is fine-tuned for 2200 iterations (approximately 20 epochs) with a learning rate of 0.0001. We utilize Google Colab GPU with a batch size of 10 to fine-tune the model.

## IV. RESULTS

In this section, we provide a comparison between our fine-tuned model and the baseline Real-ESRGAN model to highlight the improvements achieved through the fine-tuning process.

As described in the SRGAN [9] paper, the existing quantitative metrics, such as peak signal-to-noise ratio (PSNR) lack the ability to accurately reflect the perceptual preferences of human observers. These metrics often fail to evaluate intricate texture information and fine details. It is important to note that achieving high PSNR value does not necessarily guarantee superior perceptual quality. To overcome these limitations, we rely on qualitative results for a more detailed evaluation of the models. Qualitative assessment provides a more accurate and comprehensive understanding of the model's performance in underwater image super-resolution. By visually analyzing the generated images, we can assess the preservation of fine details, while also evaluating the realism and overall visual quality of the images.

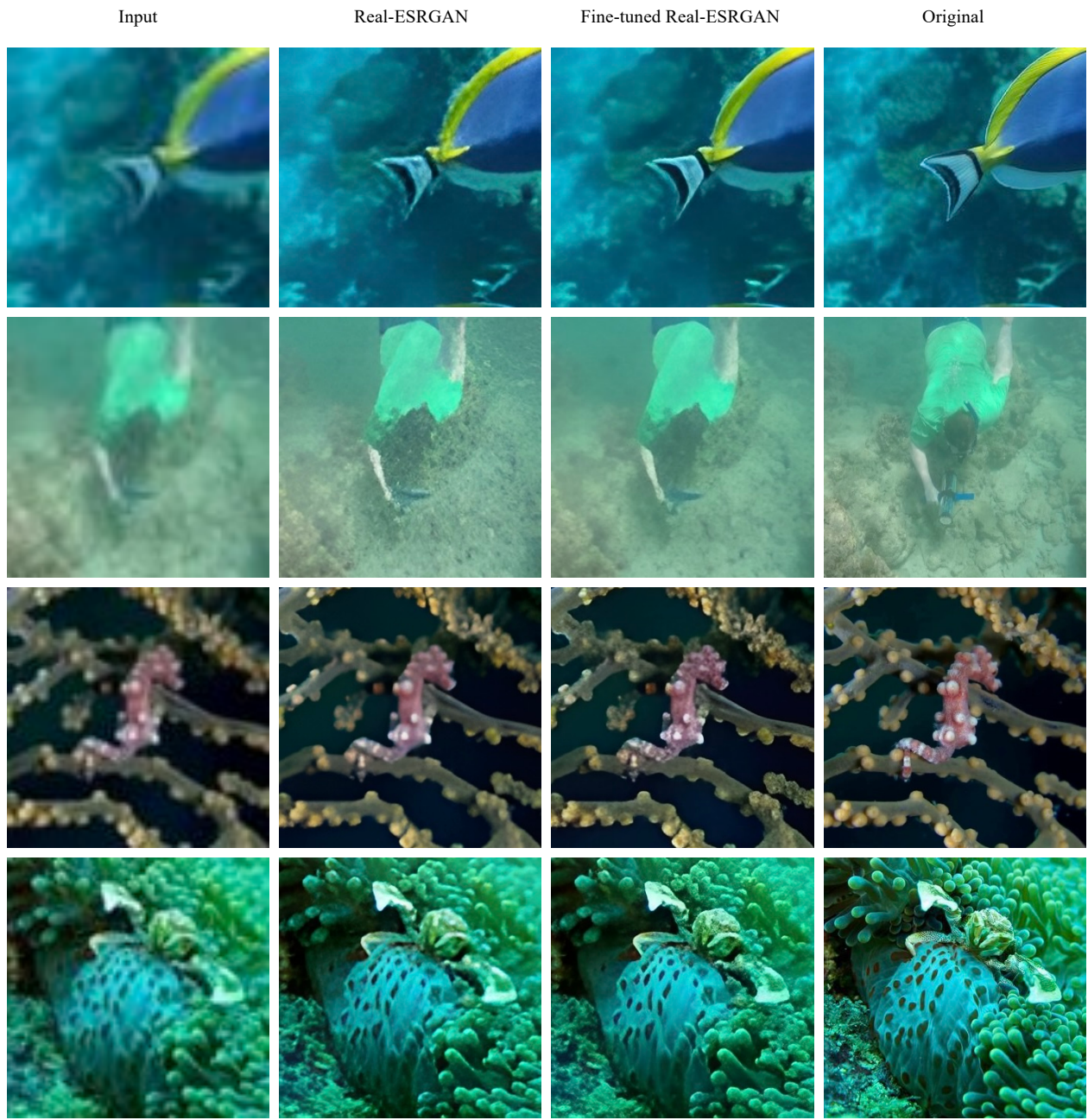

Fig. 2. Magnified regions of underwater images. Columns 1–4 are input images, Real-ESRGAN outputs, fine-tuned Real-ESRGAN outputs, and original images.

In our experiments, the resolution of the input images is increased by using a scale factor of 4×. Fig. 1 shows the results for different images from the testing folder of the USR-248 dataset. Furthermore, for a better comparison of fine details, magnified regions of underwater images have been shown in Fig. 2.

The obtained results demonstrate that the fine-tuned model outperforms the Real-ESRGAN model in terms of recovering realistic and natural textures and generating images with enhanced visual quality. Furthermore, the comparison reveals that the fine-tuned model can better preserve fine details, such as intricate textures. These findings confirm the effectiveness of our fine-tuning approach in achieving superior performance and producing higher-quality images in the context of underwater image super-resolution.

Our proposed model exhibits great potential for a wide range of real-world underwater applications. AUVs can benefit from this model in tasks like underwater mapping, object detection, and navigation, enabling more accurate data collection and analysis. In the field of marine research, our model can facilitate species identification and analysis of underwater ecosystems by improving the visual quality of underwater images. Furthermore, in underwater archaeology, underwater surveillance systems, and other related fields, the enhanced image quality provided by the model can improve decision-making and analysis in challenging underwater environments.

## V. Conclusion and future work

In this paper, we fine-tuned the pre-trained Real-ESRGAN model specifically to improve the resolution and perceptual quality of underwater images. Our proposed model outperforms the Real-ESRGAN model and achieves superior perceptual quality, making it well-suited for real-world underwater applications.

As a potential direction for future work, the A-ESRGAN model [10] can be fine-tuned to improve the visual quality of underwater images. This model incorporates a multi-scale attention U-Net discriminator and demonstrates improved performance compared to the Real-ESRGAN model.